\documentclass[10pt,twocolumn,letterpaper]{article}

\usepackage[pagenumbers]{cvpr} 

\usepackage{graphicx}
\usepackage{amssymb}
\usepackage{booktabs}
\usepackage{amsmath}
\usepackage{mathrsfs}
\usepackage{gensymb}
\usepackage{multirow}
\usepackage{algorithm} 
\usepackage{pifont}

\usepackage{amsmath,amsfonts,bm}

\def\eqref#1{equation~\ref{#1}}

\def\1{\bm{1}}

\DeclareMathAlphabet{\mathsfit}{\encodingdefault}{\sfdefault}{m}{sl}
\SetMathAlphabet{\mathsfit}{bold}{\encodingdefault}{\sfdefault}{bx}{n}

\usepackage{algorithm,algorithmic}

\usepackage{arydshln}

\newcommand{\cmark}{\ding{51}}%
\newcommand{\xmark}{\ding{55}}%
\usepackage[pagebackref,breaklinks,colorlinks]{hyperref}
\usepackage{pifont}
\usepackage{enumerate}

\usepackage[capitalize]{cleveref}
\crefname{section}{Sec.}{Secs.}
\Crefname{section}{Section}{Sections}
\Crefname{table}{Table}{Tables}
\crefname{table}{Tab.}{Tabs.}

\def\cvprPaperID{} 
\def\confName{CVPR}
\def\confYear{2023}

\begin{document}

\title{NeRFEditor: Differentiable Style Decomposition for Full 3D Scene Editing}

\author{Chunyi Sun, Yanbin Liu, Junlin Han, Stephen Gould\\
Australian National University \\
Project page: \textcolor{black}{\href{https://chuny1.github.io/NeRFEditor/nerfeditor.html}{https://chuny1.github.io/NeRFEditor/nerfeditor.html}}\\
}
\maketitle

\begin{abstract}

We present NeRFEditor, an efficient learning framework for 3D scene editing, which takes a video captured over 360$\degree$ as input and outputs a high-quality, identity-preserving stylized 3D scene. Our method supports diverse types of editing such as guided by reference images, text prompts, and user interactions. 
We achieve this by encouraging a pre-trained StyleGAN model and a NeRF model to learn from each other mutually. 
Specifically, we use a NeRF model to generate numerous image-angle pairs to train an adjustor, which can adjust the StyleGAN latent code to generate high-fidelity stylized images for any given angle.
To extrapolate editing to GAN out-of-domain views, we devise another module that is trained in a self-supervised learning manner. This module maps novel-view images to the hidden space of StyleGAN that allows StyleGAN to generate stylized images on novel views. These two modules together produce guided images in 360$\degree$ views to finetune a NeRF to make stylization effects, where a stable fine-tuning strategy is proposed to achieve this.
Experiments show that NeRFEditor outperforms prior work on benchmark and real-world scenes with better editability, fidelity, and identity preservation.

\end{abstract}

\section{Introduction}
\label{sec:intro}

Imaging if we can take a short video and generate an editable 3D scene. This ability will facilitate interesting applications in game and movie product, e.g., freely editing an identity-preserving character in real scene according to various user demands. 
Current techniques require 3D modeling expertise and long development times per scene, which is infeasible for real-time and customized editing.  

\begin{figure}[t]
     \centering
     \includegraphics[width=1\linewidth]{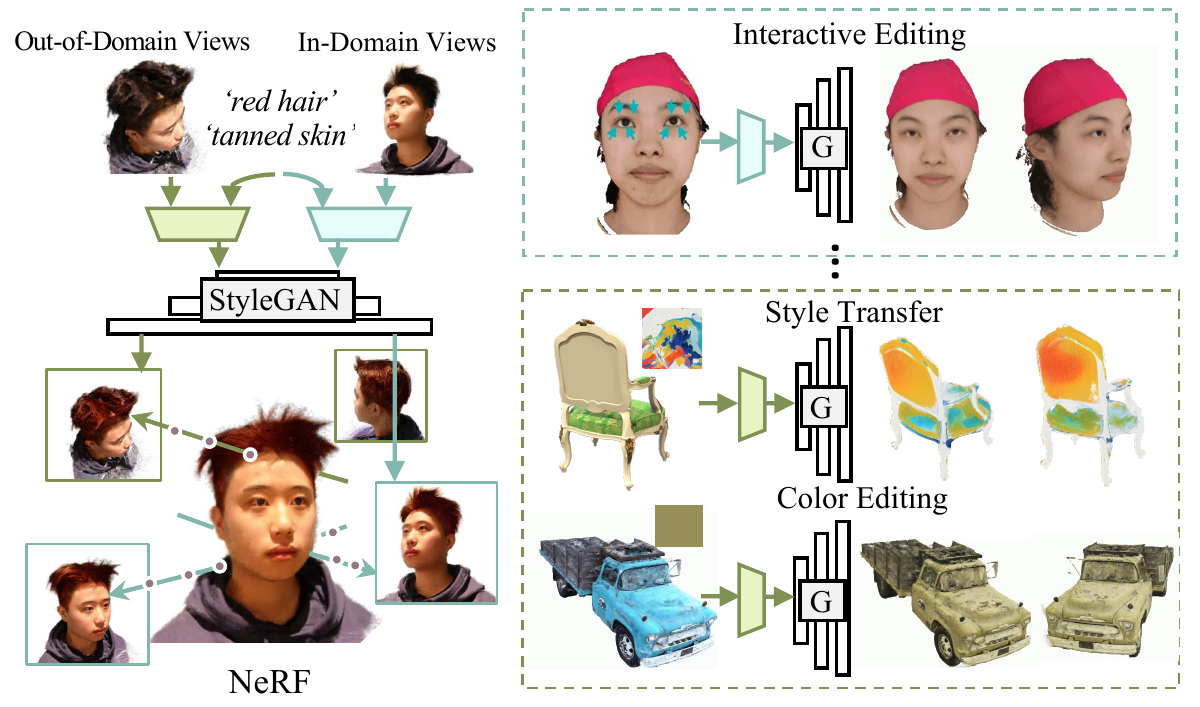}
     \caption{\label{fig:intro}
     \textbf{Method overview.} (Left) We propose two modules for a pre-trained StyleGAN to generate stylized images from in-domain views and out-of-domain views. Then, we use these styled images to finetune a conditional NeRF for 3D-consistent 360$\degree$ scene editing. (Right) Our method supports diverse types of editing.
     } 
\vspace{-0.2cm}
\end{figure}

There are a few previous works that take steps towards enabling similar abilities, as summarized in Tab.~\ref{tab:cmp_prev}.
Using existing latent space manipulation techniques, 2D GANs~\cite{GANSpace,InterpretingGAN,StyleGAN} can produce stylized images from multiple camera poses. However, these 2D methods are confined to the training pose distribution and cannot ensure 3D consistency. 
3D-aware synthesis methods~\cite{Gu2022StyleNeRFAS,EG3D,piGAN,ref:GIRAFFE} can generate multiview-consistent images using unstructured 2D images for training. Since their primary goal is not free-view editing, they experience noticeable quality degradation when extrapolating to unseen camera views.
Directly editing the neural radiance field (NeRF) is a promising direction for full scene editing. However, existing NeRF editing methods~\cite{EditNeRF,CLIPNeRF} only support basic shape and color editing of simple objects. These approaches demand diverse 3D real-scene data for training, which is expensive to obtain.

In this paper, we present \textit{NeRFEditor}, an efficient learning 
framework for 3D editing of real scenes, which supports diverse editing types to produce high-fidelity, identity-preserving scenes (Fig.~\ref{fig:intro}). 
Our framework leverages the novel-view synthesis merit of a NeRF and the well-behaved latent space of a pre-trained StyleGAN. 
The former ensures 3D-consistent scene editing, while the later facilitates flexible manipulation and high-quality generation. However, it is not straightforward to incorporate a 3D rendering model (NeRF) and a 2D generative model (StyleGAN) into a unified learning framework. 

To generate guided stylized images on any camera pose, we design two additional modules for StyleGAN. First, a \emph{latent code adjustor} is devised to take the camera parameter as input and change the image to the target view by manipulating the latent code of StyleGAN.
In the latent code adjustor, we introduce a differentiable decompositor to decomposite the pre-trained StyleGAN latent space $\mathcal{W}$ into orthogonal basis. Then, the adjustor disentangles the pose from other styles to generate multiview stylized images to guide NeRF stylization. Second, to overcome the limitation that StyleGAN cannot produce images on the out-of-domain poses, we design a novel \textit{hidden mapper}. This maps novel-view images to the hidden feature space of StyleGAN, where style mixing is applied to generate stylized images to guide the NeRF stylization training. We develop a self-supervised training algorithm for hidden mapper so that StyleGAN can adapt generative ability to out-of-domain views. 

For 3D-consistent scene editing, we use the above-generated in-domain and out-of-domain stylized images to finetune an adapted conditional NeRF model. 
The resulting model supports a wide range of 3D editing applications, such as text-prompt editing, image-guided editing, interactive editing, and style transfer. 
It outperforms prior work on standard evaluation metrics and image fidelity. 

To sum up, we make the following contributions:
\begin{itemize}
  \vspace{-5pt}
  \itemsep -3pt\partopsep -7pt
  \item We propose an efficient framework for full 3D scene editing, which only takes a few minutes on a single GPU. A summary of the features of our method over existing approaches is given in Tab.~\ref{tab:cmp_prev}.
  
  \item We design a differentiable latent code adjustor to disentangle camera pose from other styles, which brings NeRF and pre-trained StyleGAN into a unified learning framework. 

  \item  We devise a self-supervised hidden mapper to facilitate the out-of-domain style editing. 

  \item We achieve state-of-the-art results on a public benchmark dataset and a newly-collected real-scene human dataset. Our method shows promising application potentials for diverse high-quality editing types. \footnote{Code and dataset will be released.}
\end{itemize}

\begin{table}[t]
\centering
\setlength{\tabcolsep}{1.1pt}
\resizebox{1.0\linewidth}{!}{
\begin{tabular}{l|ccccc}
\hline
Method     & \shortstack{360$\degree$\\Editing} & \shortstack{w/o 3D\\Auxiliary data} & \shortstack{Real image\\Editing} & \shortstack{3D\\Consistent} & \shortstack{Real-time\\Editing}\\
\hline
2D GANs~\cite{GANSpace,InterpretingGAN,StyleGAN}             & \xmark & \cmark & \cmark & \xmark & \cmark \\
3D-aware~\cite{Gu2022StyleNeRFAS,EG3D,piGAN,ref:GIRAFFE}  & \xmark & \cmark & \xmark & \cmark & \xmark \\
NeRF editing~\cite{EditNeRF,CLIPNeRF}        & \cmark & \xmark & \xmark & \cmark & \xmark \\
Ours                 & \cmark & \cmark & \cmark & \cmark & \cmark \\
\hline
\end{tabular}}
\caption{\textbf{{Compare the main features with previous works.} 
} 
}
\label{tab:cmp_prev}
\vspace{-0.3cm}
\end{table}

\section{Related Work}
\label{sec:rw}
\begin{figure*}[t]
     \centering
     \includegraphics[width=0.8\linewidth]{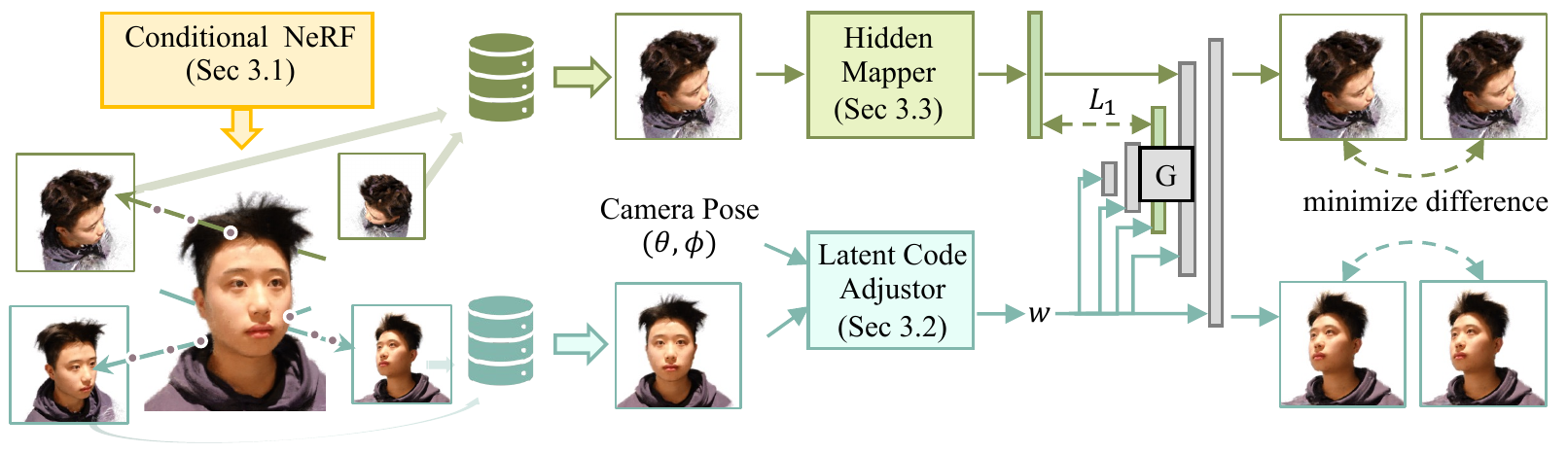}
     \caption{\label{fig:overview}\textbf{The framework of NeRFEditor.} 
     (1) The \textit{\textbf{conditional NeRF}} is pre-trained on the original scene to produce sufficient (image, pose) pairs. 
     (2) The pairs are used to train a \textit{\textbf{Latent Code Adjustor}}, which disentangles camera pose from other styles in the learned latent space of StyleGAN. 
     (3) To enable out-of-domain extrapolation, a \textit{\textbf{Hidden Mapper}} maps novel-view images to the hidden feature space of StyleGAN. 
     Finally, conditional NeRF will be finetuned for 3D-consistent editing (Sec.~\ref{sec:style_converting}). 
     } 
\end{figure*}

\textbf{2D Latent Space Manipulation.} 
Previous 2D GAN manipulation works~\cite{GANSpace, UnsupervisedDO, Voynov2020UnsupervisedDO} show that the latent space of pre-trained GANs can be decomposed to control the image generation process for attribute editing. 
The viewing direction can also be disentangled from other attributes, which allows a GAN to produce images from different viewing directions. 
However, since the training dataset cannot cover a diverse and continuous range of viewing directions, both the supervised~\cite{InterpretingGAN,Kulkarni2015DeepCI,Tran2017DisentangledRL,Singh2019FineGANUH} and unsupervised~\cite{GANSpace,UnsupervisedDO,Voynov2020UnsupervisedDO,Shen2021ClosedFormFO} manipulation methods struggle to make accurate and out-of-domain control of viewing directions. 
Moreover, these 2D methods have difficulties in generating a 3D-consistent scene. 
In contrast, we design a mutual NeRF-StyleGAN learning framework, which inherits the 3D-consistent ability of NeRF and maintains the latent space manipulation property of StyleGAN. 
Our framework can also extrapolate the editing feature outside the training viewing distribution with a novel self-supervised hidden mapper.

\textbf{3D-aware Synthesis.} 
Most recent 3D-aware synthesis methods rely on the NeRF~\cite{NeRF} technique to generate 3D-consistent images using unstructured 2D training images. 
As a pioneer work, pi-GAN~\cite{piGAN} represents the implicit neural radiance field by a SIREN network~\cite{ref:SIREN} and applies the FiLM~\cite{ref:FiLM} conditioning on SIREN, leading to the view-consistent synthesis. 
FENeRF~\cite{FENeRFFE}, EG3D~\cite{EG3D} and StyleNeRF~\cite{Gu2022StyleNeRFAS} take a two-step procedure: 1) condition a NeRF on viewpoint and style to render a low-resolution feature map; 2) convert the map to a high-resolution image with a CNN-based generator. 
Despite the view-consistent synthesis, the 3D-aware methods encounter two problems that prevent their direct application to real 3D scene editing: poor out-of-domain view extrapolation and inaccurate GAN inversion. 
In our method, the first problem is addressed by a well-devised hidden mapper that can generate out-of-domain stylized images. 
For the second problem, 2D GAN inversion has been well-explored, and our proposed differentiable adjustor further improves the GAN inversion. 

\textbf{Editable NeRF.} NeRF~\cite{NeRF} encodes the radiance field of a scene in the MLP weights to conduct novel-view rendering. 
As an implicit function, NeRF is challenging to edit. Existing works~\cite{EditNeRF,CLIPNeRF,Yuan2022NeRFEditingGE,MoFaNeRFMF} only support editing on local parts of object or with simple types. 
EditNeRF~\cite{EditNeRF} was the first work to edit the shape and color of NeRF on local parts of simple objects. 
CLIP-NeRF~\cite{CLIPNeRF} improves EditNeRF by leveraging a CLIP model to support text prompt or exemplar images, but still on simple objects. 
Compared with them, our method can edit the complex scenes and objects (e.g.,~human face) to generate high-fidelity 3D-consistent images. Meantime, we support more diverse editing types such as reference image, text, and user interactions. 

\section{Mutual NeRF-StyleGAN Training}
\label{sec:method} 
As shown in Fig.~\ref{fig:overview}, 
we design an efficient learning framework for 
360$\degree$ scene editing, which leverages the 3D consistency strength of NeRF and latent space manipulation ability of StyleGAN. 
To facilitate mutual training, we adapt a conditional NeRF (Sec.~\ref{sec:cond_nerf}) and devise two novel modules 
for StyleGAN: a latent code adjustor (Sec.~\ref{sec:latent_code_adjustor}) and a hidden mapper (Sec.~\ref{sec:hidden_mapper}). Finally, the conditional NeRF is finetuned for 3D-consistent style editing (Sec.~\ref{sec:style_converting}).

\subsection{Conditional NeRF} 
\label{sec:cond_nerf}
We start by adapt a conditional NeRF model to (1) produce (image, pose) pairs for style manipulation in StyleGAN and (2) use manipulated images to effectively finetune the NeRF model for 3D consistent scene editing (Sec.~\ref{sec:style_converting}). 

The NeRF model is a continuous function $\mathbf{F}$ that maps 3D coordinates $\mathbf{x}= (x,y,z)$, and viewing direction $\mathbf{v} = (\theta, \phi)$ to colors $\mathbf{c} = (r,g,b)$ and density $\sigma$. 
To enable the interaction with StyleGAN for scene editing, we condition NeRF on a style code $\bm{\alpha}$ and an appearance code $\bm{\beta}$. 
We assign different style codes for the original scene and stylized scene, and assign unique appearance codes for each different image in the stylized guided set. The style code allows us to train the NeRF conditioned on different styles and the appearance code could help further handle the appearance inconsistency in the stylized guided set. 
Here, $\bm{\alpha}$ and $\bm{\beta}$ are encoded by trainable MLPs $\mathcal{E}_{\text{style}}(\cdot)$ and $\mathcal{E}_{\text{app}}(\cdot)$; $\mathbf{x}$ is encoded by a hash encoder $\mathcal{H}(\cdot)$~\cite{Mller2022InstantNG}; and $\mathbf{v}$ is encoded by position encoding $\mathcal{R}(\cdot)$. The colors $\mathbf{c}$ and density $\sigma$ are predicted as follows:
\begin{align}
    \nonumber
    & \mathbf{F_{\text{dens}}}: (\mathcal{E}_{\text{style}}(\bm{\alpha})\,,\mathcal{E}_{\text{app}}(\bm{\beta}) \,,\mathcal{H}(\mathbf{x})) \mapsto (\mathbf{f}_{\text{geo}}, \sigma)\,, \\ 
    \nonumber
    & \mathbf{F}_{\text{color}}:
    (\mathcal{E}_{\text{style}}(\bm{\alpha})\,,\mathcal{E}_{\text{app}}(\bm{\beta})\,,\mathcal{R}(\mathbf{v})\,, \mathbf{f}_{\text{geo}}) \mapsto \mathbf{c}\,,
\end{align}
where $\mathbf{F_{\text{dens}}}$ and $\mathbf{F}_{\text{color}}$ denote the density and color sub-functions of $\mathbf{F}$. 

\subsection{Latent Code Adjustor} 
\label{sec:latent_code_adjustor}

\begin{figure}[h]
     \centering
     \includegraphics[width=0.97\linewidth]{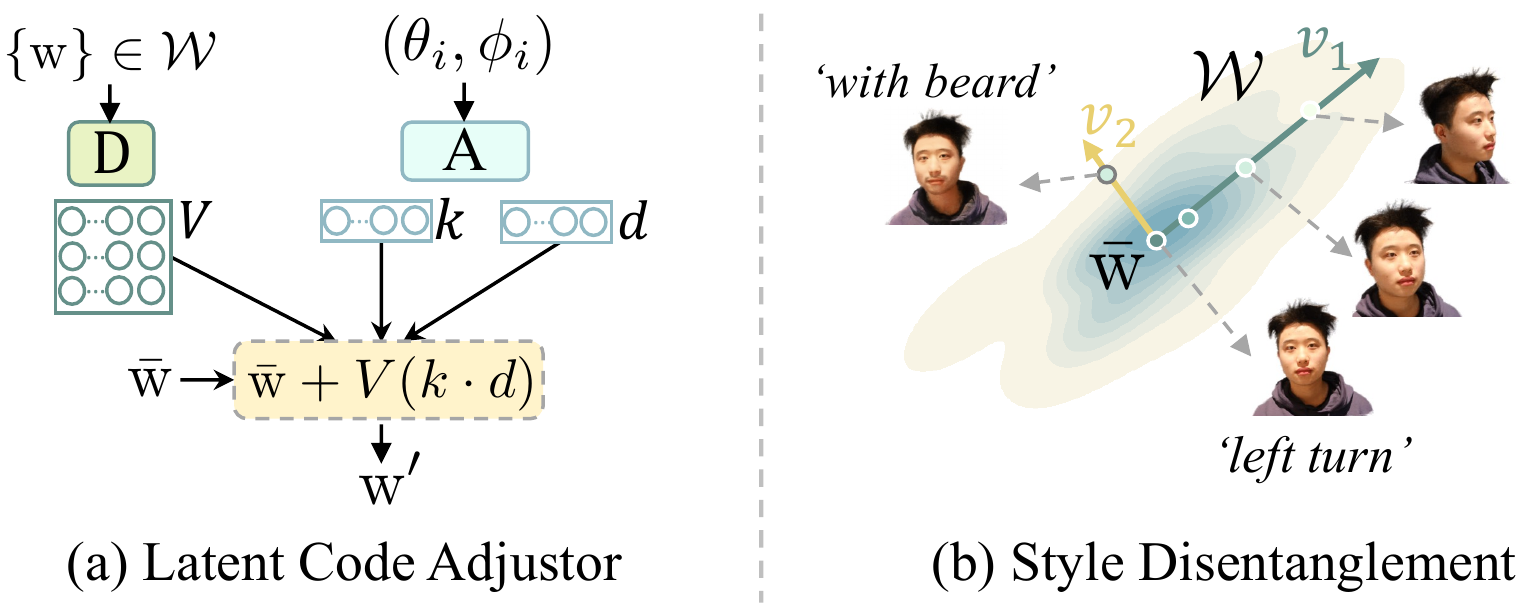}
     \vspace{-0.2cm}
     \caption{\label{fig:stage1}\textbf{Latent Code Adjustor.} (a) The latent space $\mathcal{W}$ is decomposed by a differentiable node $\mathbf{D}$ to get an orthogonal basis $V$. 
     An adjustor $\mathbf{A}$ 
     predicts the view-related coordinates $k$ of $V$ and their strengths $d$. Then, an original code $\bar{w}$ is adjusted to $w^{\prime}$. 
     (b) An example of how the adjustor works. Suppose the learned latent space $\mathcal{W}$ only contains two different styles: \textit{‘left turn’} and \textit{‘with beard’}. Hence, $\mathcal{W}$ is decomposed into two directions $v_1$ and $v_2$. Traversing over $v_1$ will only enforce pose variations.
     }  
     \vspace{-0.4cm}
\end{figure}

After obtaining sufficient (image, pose) pairs, we employ a pre-trained StyleGAN to generate high-fidelity stylized images from multiple camera views. 
Recent works~\cite{GANSpace,InterpretingGAN} found that the pre-trained StyleGAN has a well-behaved latent space, which involves interpretable styles such as pose, color, expression, etc. 
Motivated by this, we devise a differentiable decomposition module, named \textit{latent code adjustor}, to disentangle the camera pose from other styles, thereby enabling view-consistent image stylization. 
To ensure the high-fidelity, we restrict the camera pose range to lie in StyleGAN's training pose distribution~\cite{piGAN,FENeRFFE,ref:GIRAFFE}. This range is defined as the in-domain camera views $\mathbf{V}_{\text{in}}$. 

In StyleGAN, a mapping network $\mathbf{M}$ converts a vector $z \in \mathcal{Z}$ (sampled from a normal distribution) into a latent code $w \in \mathcal{W}$. The latent code $w$ is then passed to a generator $\mathbf{G}$ to generate images. To manipulate a real-world image, GAN inversion methods are used to map the image to the latent code, and most methods~\cite{Hu2022StyleTF,Dinh2022HyperInverterIS,tov2021designing} train an encoder $\mathbf{E}$ to map an image back to the latent code.
However, the inverted latent code may not generate a nearly identical image, hindering high-fidelity manipulation. 
We add an MLP layer on the top of a pre-trained encoder $\mathbf{E}$ to refine the predicted latent code for better GAN inversion. 

\textbf{Differentiable style decomposition.} 
The architecture of the latent code adjustor is shown in Fig.~\ref{fig:stage1}(a). 
A decomposition node $\mathbf{D}$ is applied on the intermediate latent space $\mathcal{W}$ to get an orthogonal basis $V$ representing different styles. 
Specifically, we first sample a large batch of $z$ from the normal distribution and use the mapping network $\mathbf{M}$ to get a large set of $w \subseteq \mathcal{W}$. Then, we compute the covariance matrix $\text{COV}(\{w\})$ and solve an eigen-decomposition problem~\cite{ref:eigen} on $\text{COV}(\{w\})$ to obtain the orthogonal basis $V$. 

The eigen-decomposition is differentiable and trained end-to-end with other modules. We explicitly derive gradients using DDNs technique~\cite{Gould2019DeepDN} to efficiently back-propagate through eigen-decomposition.\footnote{Problem formulation and derivation are in the Appendix.}

\textbf{Latent code adjustment.} Given the orthogonal basis $V$ and a latent code $w$, image editing can be done as $\small w^{\prime} = w + Vx$, where entry $x_k$ of $x$ is a separate control parameter. 

To perform target view editing, we need to determine the view-related coordinates and the corresponding adjusting strengths that need to traverse that coordinates to the target view.
Given the target pose $(\theta, \phi)$, we first use a lightweight classifier $\mathbf{C}$ to classify the view-related coordinates as $k=\text{sigmoid}(\mathbf{C}(\theta, \phi))$ and then use a lightweight regressor $\mathbf{R}$ to predict the corresponding view-adjusting strengths as $d=\mathbf{R}(\theta, \phi)$. Now, given a latent code $w$, we can generate the edited image $\widehat{I}^{(\theta,\phi)}$ of the target view by adjusting the latent code: $w^{\prime} = w + V(k\cdot d), \widehat{I}^{(\theta,\phi)} = \mathbf{G}(w^{\prime})$.

To train the latent code adjustor, we use the unedited frontal image $\bar{I}$ to obtain the latent code $\bar{w} = \mathbf{E}(\bar{I})$. With the generated target view image $\widehat{I}^{(\theta,\phi)}$ and the groundtruth $I^{(\theta,\phi)}$ generated by NeRF, we employ four loss terms as following:
\begin{equation}
    L_{total} = L_2 + L_{vgg} + L_{id} + L_{reg}\,.
\end{equation}
Here, the first three terms represent the $\ell_2$ distance, VGG perceptual distance~\cite{ref:perceptual}, and identity distance computed between $\widehat{I}^{(\theta,\phi)}$ and $I^{(\theta,\phi)}$, respectively. 
The last term $L_{reg}$ is an entropy regularization term to encourage $k$ to be sparse, as only few coordinates are view-related. 

\textbf{In-domain Stylized Set $\mathcal{I}_{in}$.} 
We can apply any existing latent code editing methods~\cite{GANSpace,InterpretingGAN} on the frontal latent code $\bar{w}$ to get $w_{style}$. We use the latent code adjustor to achieve generating multi-view stylized images. 
To generate the in-domain stylzied set, we generate the stylized image for every camera pose in the NeRF-guided image set.

\subsection{Hidden Mapper}
\label{sec:hidden_mapper}

\begin{figure}[t]
     \centering
     \includegraphics[width=1\linewidth]{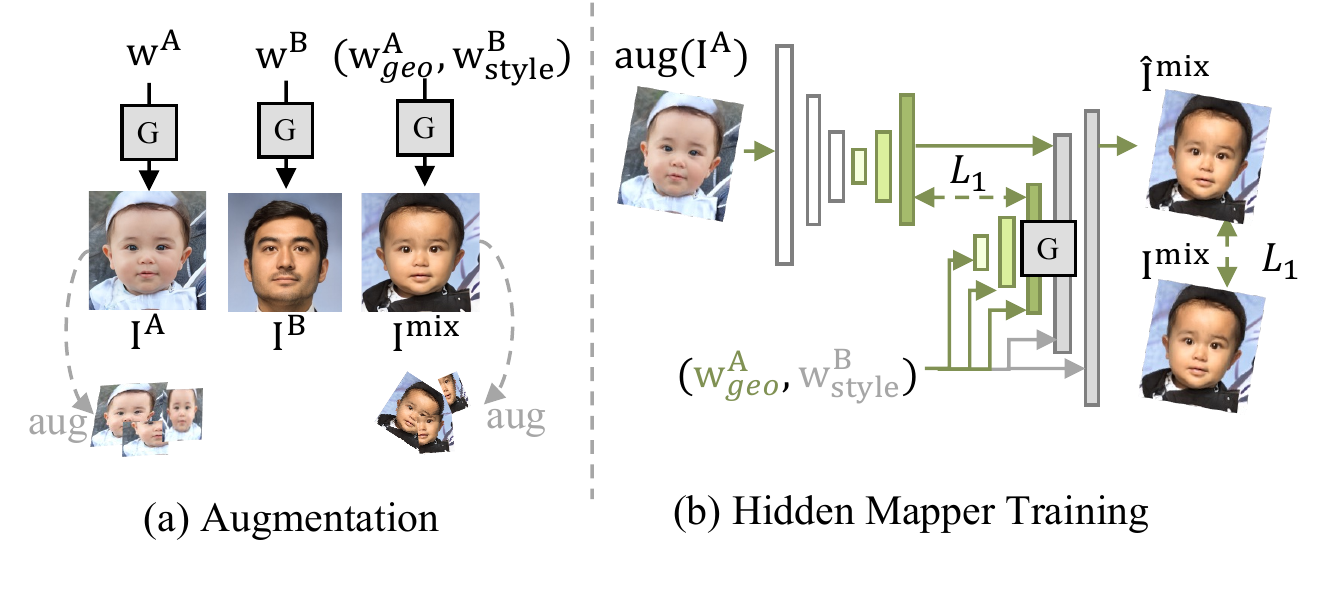}
     \vspace{-0.6cm}
     \caption{\textbf{Hidden Mapper.} (a) The codes are split as a geometric code $w_{geo}$ and a style code $w_{style}$. Same augmentation is applied on hidden features controlled by $w_{geo}$ and style-mixed images controlled by mixed codes $(w_{geo}^A, w_{style}^B)$. 
     (b) After augmentation, Hidden Mapper is trained with the $L_1$ reconstruction loss on both hidden space and image space.} 
     \label{fig:hp}
\vspace{-0.2cm}
\end{figure}

Despite the high-fidelity generation in the training pose domain, StyleGAN struggle to extrapolate to extreme camera views outside the pose distribution, denoted as $\mathbf{V}_{\text{out}}$. 
We design a \textit{hidden mapper} $\mathbf{H}$ to tackle a limitation of StyleGAN (extrapolation inability of unseen poses), thus enabling 360$\degree$ style editing.

\textbf{Self-supervised Training.} 
Latent codes of shallow layers generally control the geometric aspects (e.g. pose and shape), while latent codes of deep layers control the style attributes (e.g. skin color). Style-mixing~\cite{StyleGAN} can mix the style of two images by combining two latent codes. 
We sequentially split latent codes as $(w_{geo}, w_{style})$.
Each block of StyleGAN takes as input a hidden feature map produced by the previous block and an additional latent code.
Therefore, we can also perform style-mixing by mixing the hidden maps with $w_{style}$.
Most importantly, we observe that, when we perform augmentation operations (crop, rotate, re-scale) to the hidden map, the style can still transfer to the correct location. If we can learn a mapper that maps images to the hidden space of StyleGAN, we will be able to transfer styles to images captured on any camera location. We introduce a self-supervised algorithm to achieve this goal.

\begin{algorithm}[t]
\caption{\label{alg:Hidden}Self-supervised Hidden Mapper Training.} 
\begin{algorithmic}
\STATE \textbf{Input: }$\mathbf{G}_{geo}, \mathbf{G}_{style}$, image $I^A$, styles $w^A, w^B$ 
\STATE $F^{A}$ = aug($\mathbf{G}_{geo}(w^{A}_{geo})$) \; \# augment the hidden map 
\STATE $\widehat{F}^{A}$ = $\mathbf{H}(\text{aug}(I^{A}))$ \;\;\;\;\;\quad \# predict the hidden map
\STATE $I^{mix}$ = $\mathbf{G}_{style}(F^A,\, w^{B}_{style}$) \;\;\; \# generate mixed image
\STATE $\widehat{I}^{mix}$ = $\mathbf{G}_{style} (\widehat{F}^A,\, w^{B}_{style}$) \;\;\; \# predict mixed image 
\STATE \textbf{Return:} loss = $L_1(I^{mix},\widehat{I}^{mix}) + L_1(F^{A}, \widehat{F}^{A})$
\end{algorithmic}
\end{algorithm}

We represent the early layers of the generator that takes $w_{geo}$ as $\mathbf{G}_{geo}$ and the deeper layers of the generator that takes $w_{style}$ as $\mathbf{G}_{style}$. The generation process is denoted as $F = \mathbf{G}_{geo}(w_{geo}), I = \mathbf{G}_{style} (F,\, w_{style})$. The hidden mapper is trained to map real images to the output space of $\mathbf{G}_{geo}$, which is achieved by reconstructing both the hidden feature $F$ and the generated image $I$.  
The training algorithm is shown in Alg.~\ref{alg:Hidden} and Fig.~\ref{fig:hp}. 

\textbf{Out-of-domain stylized set $\mathcal{I}_{out}$.} 
We first get the unedited images from out-of-domain views $\mathbf{V}_{\text{out}}$ with conditional NeRF. Then hidden mapper converts these images to the hidden map $F$, which are combined with target editing styles $w_{style}$ to obtain the stylized images. 

\subsection{3D-Consistent Style Editing}
\label{sec:style_converting}

Once we have trained the latent code adjustor and the hidden mapper, the model is capable of producing stylized images of 360$\degree$. We use the images $\mathcal{I}_{ori}$ generated by original scene, in-domain set $\mathcal{I}_{in}$ and out-of-domain stylized set $\mathcal{I}_{out}$ to finetune the conditional NeRF for 3D-consistent style editing. 

To finetune conditional NeRF, we assign $\boldsymbol{\alpha}=1$ for the stylized sets and a different $\boldsymbol{\beta} \in [0,1]^{d_{\beta}}$ for each stylized image in $\mathcal{I}_{in}$ and $\mathcal{I}_{out}$. 
To focus on editing the objects, we calculate the foreground mask $M$. Then we define the following masked-guided losses \textit{w.r.t} colors $\mathbf{c}$ and density $\mathbf{\sigma}$: 
\begin{align}
    \label{eq:style_loss}
    \centering 
     & \small L_{style}  = \sum_{i,j} {{(\widehat{\mathbf{c}}_{i,j}^{style} \cdot M_{i,j} - \mathbf{c}_{i,j}^{style} \cdot  M_{i,j})}^{2}}, \\
     & \small L_{br} = \sum_{i,j} {{({\widehat{\mathbf{\sigma}}}_{i,j}^{style} \cdot  (1-M_{i,j}) - {\sigma}_{i,j}^{ori} \cdot  (1 - M_{i,j}))}^{2}}, \\
     & \small L_{ori} = \sum_{i,j} {(\widehat{\mathbf{\sigma}}_{i,j}^{ori}-{\mathbf{\sigma}}_{i,j}^{ori})^{2}},  
\vspace{-0.5cm}
\end{align}
where $i,j$ index the image, $\widehat{\mathbf{c}}$ and $\widehat{\mathbf{\sigma}}$ denote the rendered colors and density, $^{style}$ and $^{ori}$ denote the stylized and original sets. 
$L_{style}$ optimises the foreground style editing process. 
$L_{br}$ is the background regularization to constrain the density change in the background region.
$L_{ori}$ is applied to avoid forgetting the original scene, which can stabilize the training and ensure the success of $L_{br}$. 

To further handle the inconsistency in the guided training set, we follow~\cite{Niemeyer2022RegNeRFRN} to render the depth map $\mathbf{d}$ and adopt a depth regularisation term to smooth the stylized scene:  
\begin{equation}\label{eq:depth_loss}
    L_{depth} = \sum_{i,j} (\widehat{\mathbf{d}}_{i,j} - \widehat{\mathbf{d}}_{i,j+1})^{2}+(\widehat{\mathbf{d}}_{i,j} - \widehat{\mathbf{d}}_{i+1,j})^{2}\,. 
\end{equation}
The final loss used for our 3D-consistent NeRF editing is:
\begin{equation}\label{eq:nerf_loss}
    L = (L_{style} + L_{br} + L_{ori}) + \lambda \cdot L_{depth}\,.
\end{equation}

\section{Experiments}
\textbf{Datasets.} 
We evaluate \textit{NeRFEditor} on two datasets: FaceScape~\cite{MoFaNeRFMF,ref:Facescape} as a high-quality 3D face benchmark and TIFace (Tiny-scale Indoor Face collection) as a newly-collected real-world dataset.
FaceScape contains 359 different subjects each with 20 expressions, and 120 multiview images for each expression. It is captured in a lab environment and the subject wears a turban. 
To evaluate the editing capability for real 3D scenes, we collect TIFace, a real-world dataset including 10 different persons from a realistic environment, without any restriction on dressing. 
Details of the dataset collection can be found in the Appendix. 

\textbf{Evaluation Metrics.} 
Following previous works~\cite{EG3D,Gu2022StyleNeRFAS,piGAN}, 
we use various metrics to evaluate the image quality: PSNR, SSIM, and  LPIPS~\cite{Zhang2018TheUE}.
To quantify the 3D-consistency, we employ Pose (pose adjusting error)~\cite{EG3D}. 
To evaluate the identity-preservation, we use ID Score~\cite{Parkhi2015DeepFR}, which measures the confidence of classifying one image to have the same identity as the groundtruth image. 
Meanwhile, we use APS (attribute-preservation score) to measure the preservation of non-edited attributes after editing.

To compare with the state-of-the-art 3D editing method (i.e. CLIP-NeRF~\cite{CLIPNeRF}), we also report FID and FR (face recognition score) before and after editing. FR is the score of an image being recognized as human face. 

\textbf{Optimization Time Measure.}
We measure the training speed of our NeRF model in a single RTX3070 GPU, and the training time of each step is $\sim$8ms. 
During training in the original scene, 
we set the iterations to 10k for FaceSpace and 20k for TIFace. 
During finetuning, we set the iterations to 1/4 of the training iterations of the original scene. 
Thus, the total NeRF training time is less than 4 minutes for the complex scene, and 2 minutes for the simple scene. 
To render a $1024\times 1024$ image, the speed is 50 fps, which can be rendered in real time. 
The time to train the latent code adjustor and finetune the StyleGAN is around 4 minutes consistently for all datasets in $1024\times 1024$ resolution.

\subsection{Compare with Generative Baselines} 
We first compare with the 2D manipulation and 3D-aware baselines to demonstrate that our method can obtain high-fidelity and identity-preserving image generation, while also ensures good 3D-consistency after editing. 

For 2D manipulation baselines, we choose GANSpace \cite{GANSpace} and InterFaceGAN~\cite{InterpretingGAN}, both of which are able to control the pose direction. To endow them with accurate angle adjustment ability, we augment these two methods with an extra MLP to predict the angle distance to be adjusted.  
For 3D-aware baselines, we adopt two state-of-the-art methods: 
StyleNeRF~\cite{Gu2022StyleNeRFAS} and EG3D~\cite{EG3D}. 
For our method, we report on two variants: the model without 3D-consistent style editing, i.e. Ours (w/o 3D) and the whole model, i.e. Ours. For a fair comparison, we apply PTI~\cite{Roich2022PivotalTF} to all methods to achieve better inversion.

The quantitative comparison results are shown in Tab.~\ref{tab:baseline}. 
\textbf{(1)} We investigate how well each method can control the camera pose. 
To realize this, we restrict the pose range to StyleGAN's training pose domain and align the images on FaceScape.
After that, we apply each method to generate images on the same views as the testing set. This is done without editing the latent code, denoted as w/o Editing. 
We measure the ID score, PSNR, SSIM and LPIPS between the generated images and the groundtruth images. 
Tab.~\ref{tab:baseline} demonstrates that our method outperforms all 2D and 3D-aware baselines, even without 3D-consistent module (Ours (w/o 3D)). 
\textbf{(2)} Then, we measure the disentanglement (APS) and 3D-consistency (Pose) capability. 
For GANSpace and InterFaceGAN, we use their own stylization method for editing. 
For StyleNeRF and EG3D, we apply 2D editing method~\cite{GANSpace} on the frontal image and get an inverted latent code. 
According to Tab.~\ref{tab:baseline}, our method achieves the best APS score and least Pose error. The former shows that our method can better decompose edited attributes from non-edited. The latter indicates that our method can perform pose-dependent editing to ensure 3D-consistency. 

\begin{figure}[t]
     \centering
     \includegraphics[width=1\linewidth]{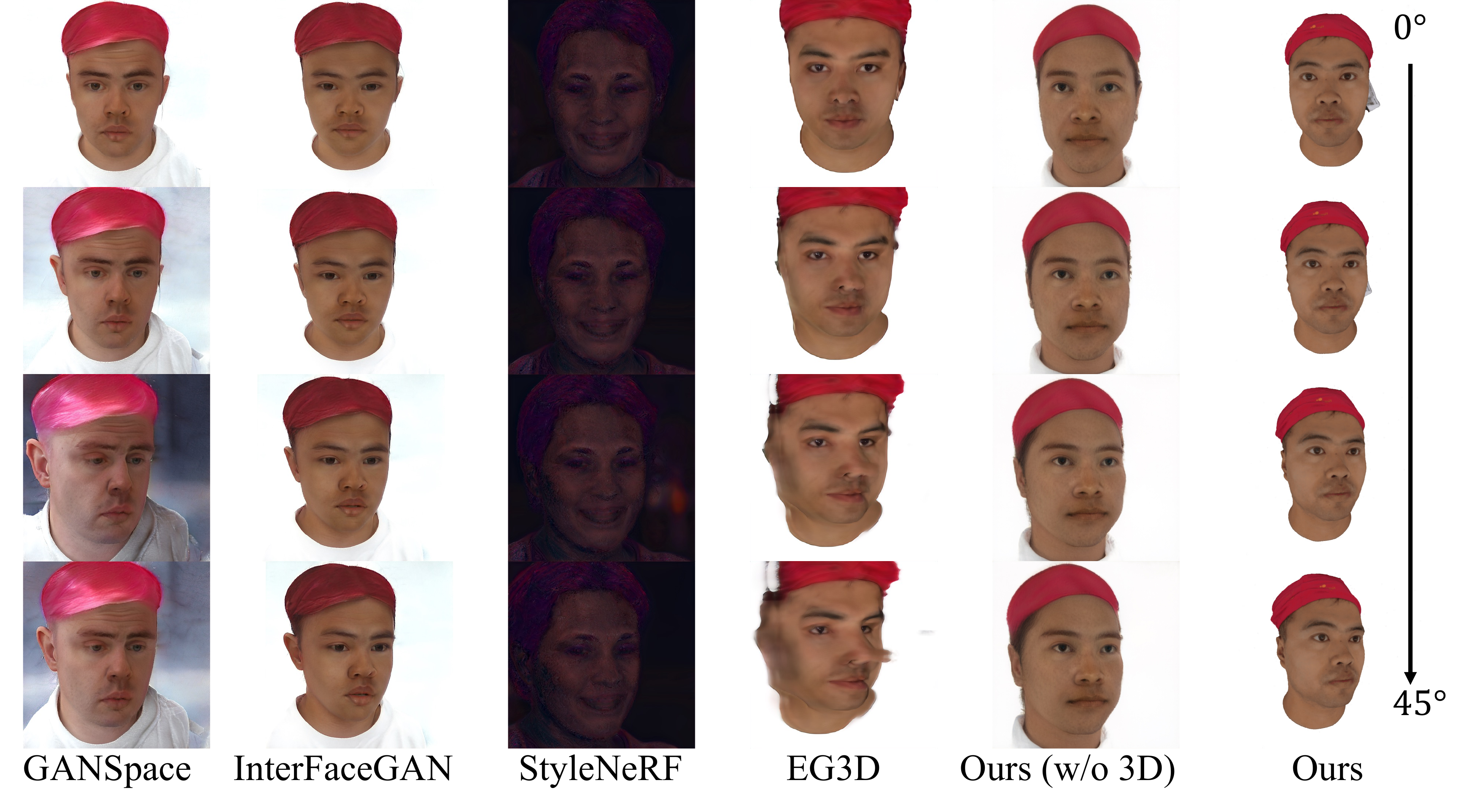}
     \caption{\textbf{Qualitative comparison with generative baselines.} The same human face is inverted by different methods to the latent code for pose manipulation.
     }
     \label{fig:baseline}
     \vspace{-0.3cm}
\end{figure}

The qualitative comparisons are shown in Fig.~\ref{fig:baseline}. 
It illustrates that all methods except for StyleNeRF can produce relatively good-quality frontal image, with the corresponding inversion strategies. However, they exhibit worse identity-preserving effect compared with our method, which is consistent with their lower ID scores in Tab.~\ref{tab:baseline}. Furthermore, when we vary the pose, the baselines degrade quickly: GANSpace incurs obvious background; InterFaceGAN has a large shift; EG3D obtains blurry results. This reveals bad disentanglement between pose and other styles. \textit{Ours (w/o 3D)} shows less accurate pose control than \textit{Ours}, verifying the necessity of the latent code adjustor.

\begin{table}[t]
\caption{\textbf{Comparison with generative baselines on FaceScape.}} 
\centering
\setlength{\tabcolsep}{2.8pt}
\resizebox{1.0\linewidth}{!}{
\begin{tabular}{l|cccc:ccc}
\hline
    &\multicolumn{4}{c}{w/o Editing}&\multicolumn{2}{c}{w Editing}\\
    & ID$\uparrow$ & PSNR$\uparrow$ & SSIM$\uparrow$ & LPIPS$\downarrow$& APS$\uparrow$ & Pose $\downarrow$\\
\hline
GANSpace~\cite{GANSpace}            &44.47 &27.75 &0.3993 &0.3848 &0.8142 &9.38 \\
InterFaceGAN~\cite{InterpretingGAN} &62.16 &29.61 &0.7623 &0.2028 &0.8592 &8.48 \\
StyleNeRF~\cite{Gu2022StyleNeRFAS}  &15.39 &27.44 &0.3482 &0.4845 &0.3421 &30.4 \\
EG3D~\cite{EG3D}                    &43.14 &29.86 &0.7738 &0.2369 &0.7694 &7.95 \\
Ours (w/o 3D)                       &80.45 &32.22 &0.8414 &0.1316 &0.8840 &8.42 \\
Ours                                &\textbf{94.59} &\textbf{33.71} &\textbf{0.8464} &\textbf{0.1277} &\textbf{0.9321} &\textbf{4.48} \\ 
\hline
\end{tabular}
}
\label{tab:baseline}
\end{table}

\begin{table*}[t]
\caption{\textbf{Compare with state-of-the-art 3D editing method.}}
\setlength{\tabcolsep}{2.0pt}
\centering
\resizebox{0.85\linewidth}{!}{
\begin{tabular}{l|cccc:cc|cc:cc}
\hline
&\multicolumn{6}{c}{FaceScape}&\multicolumn{4}{c}{TIFace}\\
\hline
&\multicolumn{2}{c}{ FID$\downarrow$ }&\multicolumn{2}{c}{ FR$\uparrow$ }& APS & ID  &\multicolumn{2}{c}{ FR$\uparrow$ }& APS & ID \\
& Before & After & Before & After&&&Before &  After &\\
\hline
CLIP-NeRF~\cite{CLIPNeRF} (color) & 6.24 & 65.38 \small{(+59.14)} & 85.72 & 69.23 \small{(-16.49)} & 77.42 & 76.38 & 88.44 & 69.32 \small{(-19.12)} & 80.96 & 75.79\\ 
Ours (color) & 6.23 & 38.31 \small{(+32.08)} & 85.43 & 81.08 \small{(-4.35)} & 86.42 & 87.42 & 87.83 & 85.73\small{(-2.10)}&87.96&89.03 \\
Ours (shape) & 6.23 & 27.31 \small{(+21.08)} & 85.43 & 81.97 \small{(-3.46)} & 85.43 & 88.53 & 
87.83&86.01\small{(-1.82)}&89.54&87.03\\
Ours (shape+color) & 6.23 & 39.42 \small{(+33.19)} & 85.43 & 81.34 \small{(-4.09)} & 84.37 & 86.98 & 87.83&84.95\small{(-2.88)}&87.04&85.96\\
\hline
\multicolumn{11}{l}{$*$ According to \cite{CLIPNeRF}, CLIP-NeRF fails to get satisfying shape editing results on the complex scene (also see Fig.~\ref{fig:cmp_sota}).}
\end{tabular}
}
\label{tab:edit_quality}
\end{table*}

\subsection{Compare with State-of-the-art 3D Editing}

We then compare with the state-of-the-art 3D editing method to demonstrate that our method supports high-fidelity editing on complex scenes. 
For 3D editing, CLIP-NeRF~\cite{CLIPNeRF} improves the previous EditNeRF~\cite{EditNeRF} and becomes the state-of-the-art method. 
On complex scenes, CLIP-NeRF supports color editing with text prompts but fails to perform satisfying shape editing~\cite{CLIPNeRF}.
So we pre-define a set of color-editing texts \emph{w.r.t} the attributes of the FaceScape dataset, such as tanned skin, red lips, and blue eyes. 
In our method, we also define a set of shape-editing texts to verify the broader editing capability.
For a fair comparison, we adapt CLIP-NeRF to use the same Hash encoding and the same configurations of the density net and color net as our method. 

\begin{figure}[!t]
     \centering
     \includegraphics[width=1\linewidth]{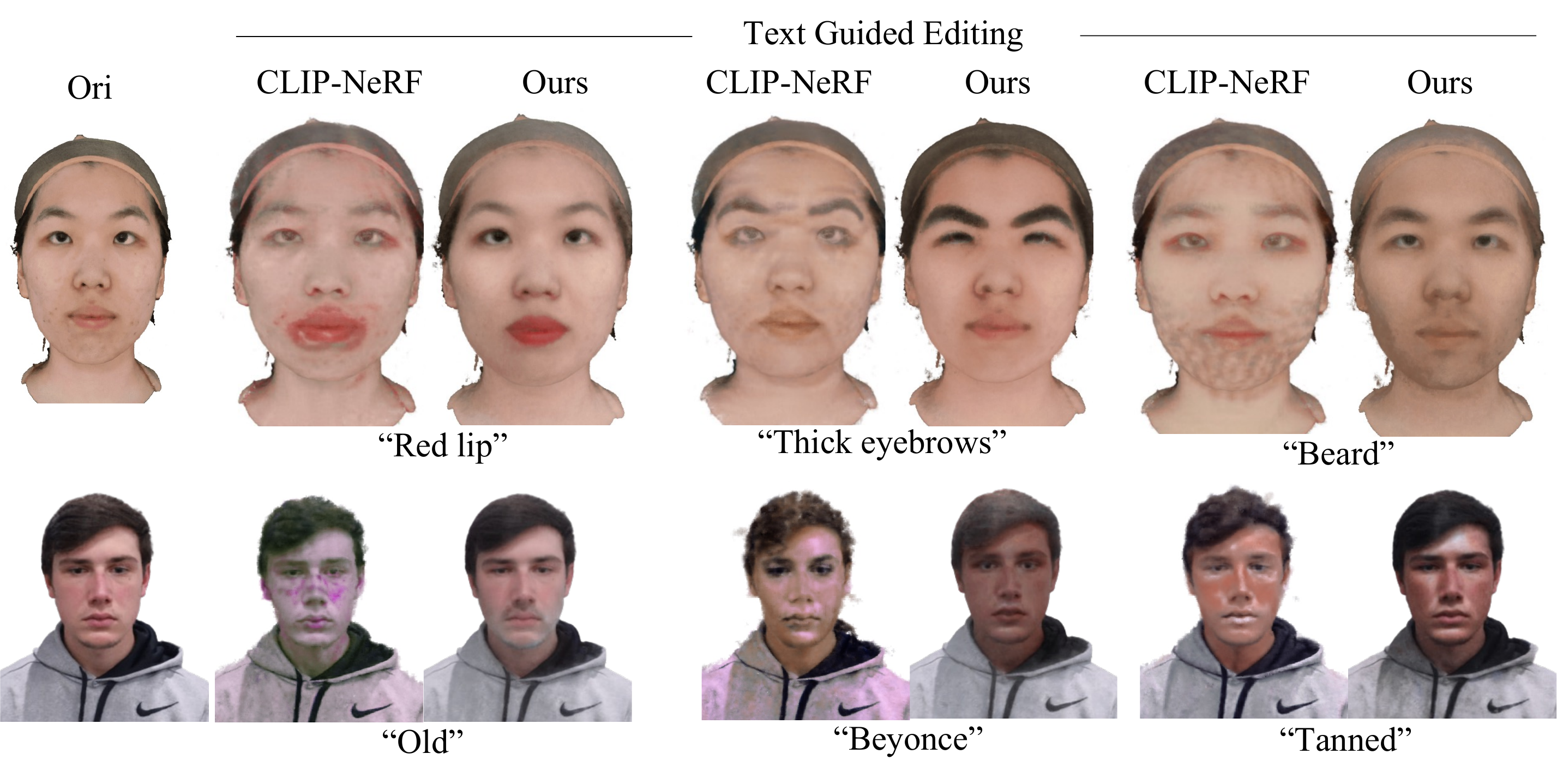}
     \caption{\textbf{Compared to CLIP-NeRF.} Our editing correctly edits the scene with given texts and maintains the quality of the original scenes. We invite readers to check the project page for better comparison.}
     \label{fig:cmp_sota}
\vspace{-0.2cm}
\end{figure}

\begin{figure}[t]
     \centering
     \includegraphics[width=1\linewidth]{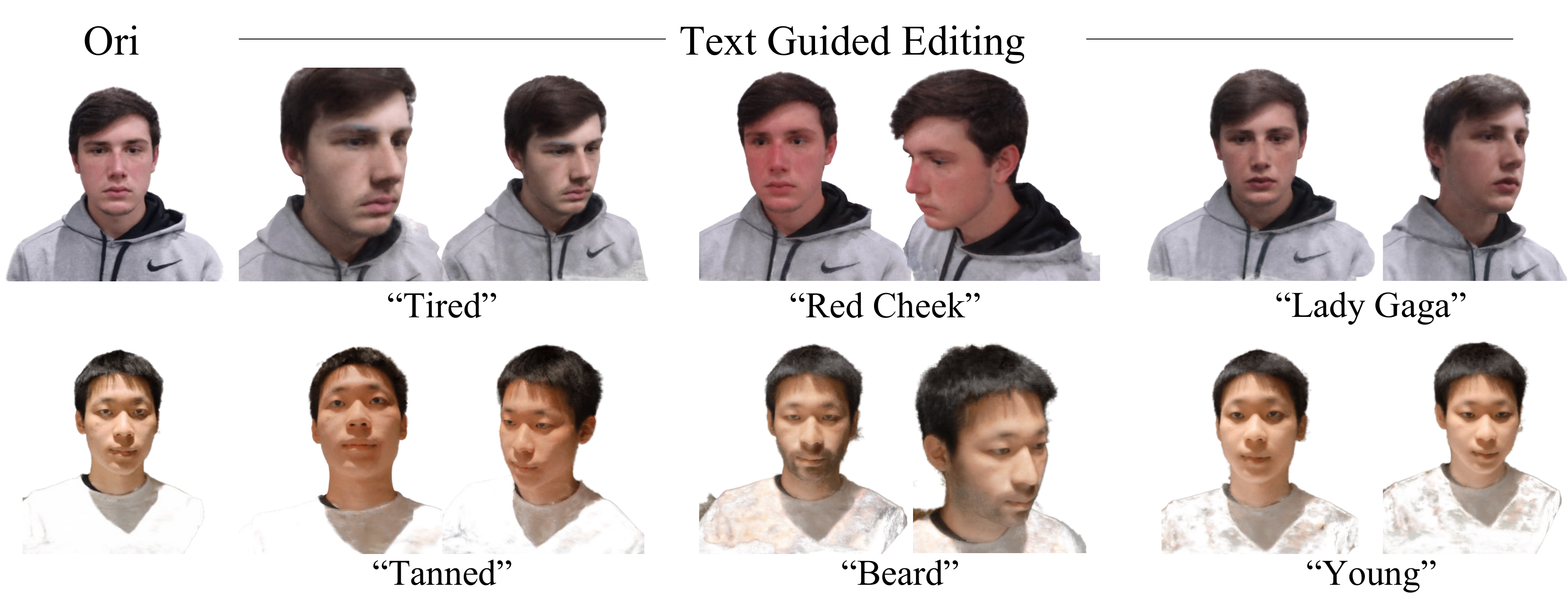}
     \caption{\textbf{Results of our method on TIFace dataset.} Our method can make accurate editing with given guided text descriptions. Please refer to our project page to see high-quality editing result with a moving camera.
     }
     \label{fig:real}
\vspace{-0.2cm}
\end{figure}

The comparison results are shown in Tab.~\ref{tab:edit_quality}. After applying color editings to FaceScape, the FID score of CLIP-NeRF increases significantly, while our method shows a reasonable shift. Similarly, after editing, the FR of CLIP-NeRF drops by 16.49, but our method only reduces by 4.35. 
For the real-world TIFace dataset, the FR differences are more severe (19.12 versus 2.10). 
The change of two metrics indicates that color editing by CLIP-NeRF significantly harms the image quality, while our method can better maintain the image quality. 
When extended to shape editing and shape+color editing, our method remains consistent performances, demonstrating the wider and more stable 3D editing capability. 
We also measure the APS and ID score to evaluate the non-edited attribute-preserving and identity-preserving features of the 3D editing methods. Our method outperforms CLIP-NeRF on both metrics.

\begin{table}[t]
\caption{\textbf{Statics from our user study.} Reported is percentage of workers voting for our method over the competing method.}
\setlength{\tabcolsep}{2pt}
\centering
\scalebox{0.85}{
\begin{tabular}{l|cc}
\hline
& Realistic$\uparrow$ & Editing Accuracy$\uparrow$\\
\hline
Ours vs. GANSpace & 99.2 & 80.0 \\
Ours vs. InterFaceGAN & 97.6 & 75.2 \\
Ours vs. StyleNeRF & 99.6 & 100.0\\
Ours vs. EG3D & 97.2 & 96.8 \\
Ours vs. CLIP-NeRF & 98.0 & 94.4 \\
\hline
\end{tabular}
}
\label{tab:user_study}
\end{table}

When applying our method to the real-world dataset, we model each scene as an unbounded scene.  We follow \cite{Zhang2020NeRFAA} to use an additional background network to encode the background, which can successfully surpass the floating artefacts and improve the image quality.
Fig.~\ref{fig:cmp_sota} shows the qualitative comparison between our method and CLIP-NeRF. 
Our method performs accurate editings that fit the text descriptions, and also maintains the image quality, 3D-consistency and identity. 
For the CLIP-NeRF, although some editings can change the image to reflect the text prompts, obvious artifacts and floating noisy pixels appear around the face region. 
In Fig.~\ref{fig:real}, we demonstrate more editing results guided by various text prompts on the real-world TIFace dataset. Our method achieves high-quality editing results and scales well for real-world scenes.

\subsection{User Study}

We also conduct a user study to compare the human evaluations between our method and all baselines and 3D editing methods. 
We randomly sample 60 scenes from the FaceScape test set and sample 5 different text descriptions for each scene. To compare with the CLIP-NeRF, we generate 360$\degree$ rotated videos before and after editing for both methods. Other methods can not extrapolate well to the extreme camera views. Thus, we generate rotated videos only (-90$\degree$, 90$\degree$) around the frontal face before and after editing for each method. 
Afterward, we use Amazon Mechanical Turk to perform a user study. We require workers to choose the video that (1) better matches the text description and (2) has a higher visual quality (i.e. fewer artifacts, clear boundaries, and more natural). 
For each pair, we shuffle the positions randomly and ask five workers to make comparison. 
Tab.~\ref{tab:user_study} summarizes the comparison, where most users consistently rate our method as the high-quality one.

\subsection{Ablation Study}
In this section, we provide ablation studies on FaceScape to verify the effectiveness of the proposed modules, training techniques, and loss terms. 

\begin{figure}[t]
     \centering
     \includegraphics[width=1\linewidth]{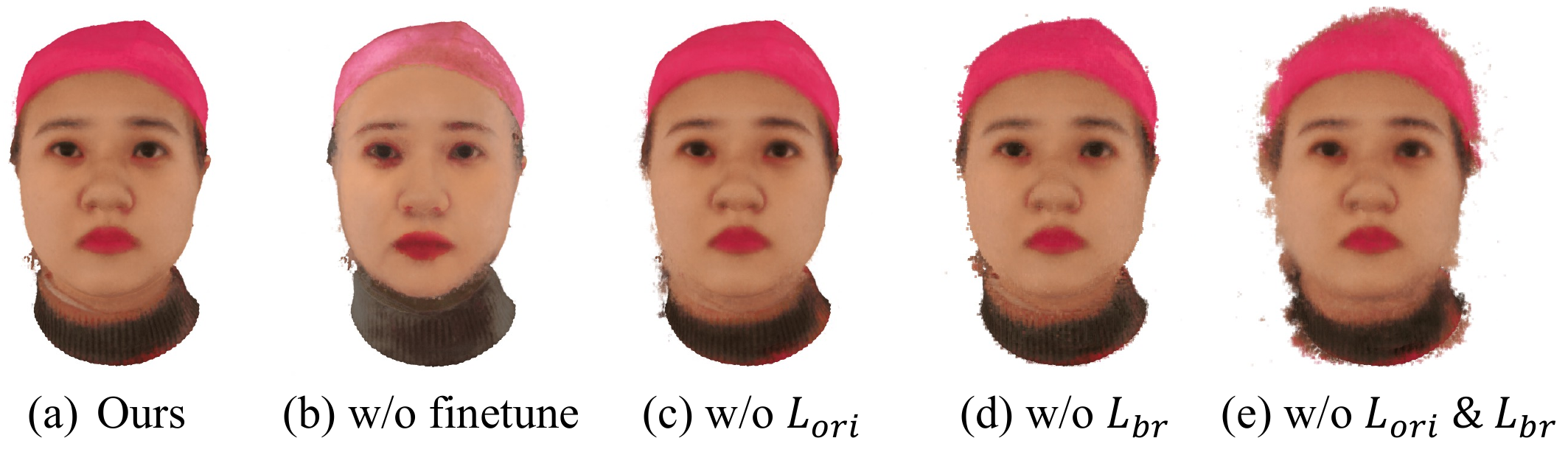}
     \vspace{-0.7cm}
     \caption{\textbf{Visual comparison of ablation study.}}
     \label{fig:albreg}
\vspace{-0.2cm}
\end{figure}

Finetuning and differentiable decomposition play important roles in the latent code adjustor. 
As shown in Tab.~\ref{tab:abl_reg}, without finetuning applied, all metrics drop significantly, especially the ID score. 
Fig.~\ref{fig:albreg}(b) also shows how finetuning could affect identity preservation. Differentiable decomposition could help the latent code adjustor gain more disentangling results, which reduces the visual degradation from the 3D-inconsistent guidance. Without differentiable approach, all metrics get worse. 

\begin{table}[t]
\caption{\textbf{Ablation study.}}
\setlength{\tabcolsep}{2pt}
\centering
\scalebox{0.9}{
\begin{tabular}{cl|cccc}
\hline
& removed module & {FID$\downarrow$} & {FR$\uparrow$} &APS & ID \\
\hline
Ours &                        &39.42  &81.34  &84.37  &86.98\\ \hline
\multirow{2}{*}{\shortstack{latent code\\adjustor}} & \textit{w/o} finetuning       &43.94  &79.40  &80.46  &79.03\\ 
    & \textit{w/o} differentiable &41.53  &80.05  &82.05  &85.49\\ 
\hline
\multirow{4}{*}{\shortstack{3D-consistent\\style editing}} & \textit{w/o} $L_{ori}$        &42.96  &80.10  &79.95  &85.69\\ 
& \textit{w/o} $L_{br}$         &42.35  &79.94  &79.06  &84.93\\
& \textit{w/o} $L_{ori} \& L_{br}$ &45.96  &76.46  &77.03  &81.94\\ 
& \textit{w/o} $L_{depth}$      &40.00  &80.97  &84.38  &86.05\\ 
\hline
\end{tabular}
}
\label{tab:abl_reg}
\end{table}

In 3D-consistent style editing, all loss terms in Eqn.~\ref{eq:nerf_loss} contribute significantly to the result (Tab.~\ref{tab:abl_reg}). 
Without $L_{br}$, we will lose the constraint to the unedited region. As shown in Fig.~\ref{fig:albreg}(d), there are obvious floating artifacts around the face boundary.  
From Fig.~\ref{fig:albreg}(c), if we remove the $L_{ori}$, the constrain ability for $L_{br}$ will be decreased, since the model tends to forget the original scene. 
Removing both $L_{br}$ and $L_{ori}$, the finetuning will fail and the scene will become diverged, thus resulting in poor visual results (Fig.~\ref{fig:albreg}(e)). 

\begin{figure}[t]
     \centering
     \includegraphics[width=1\linewidth]{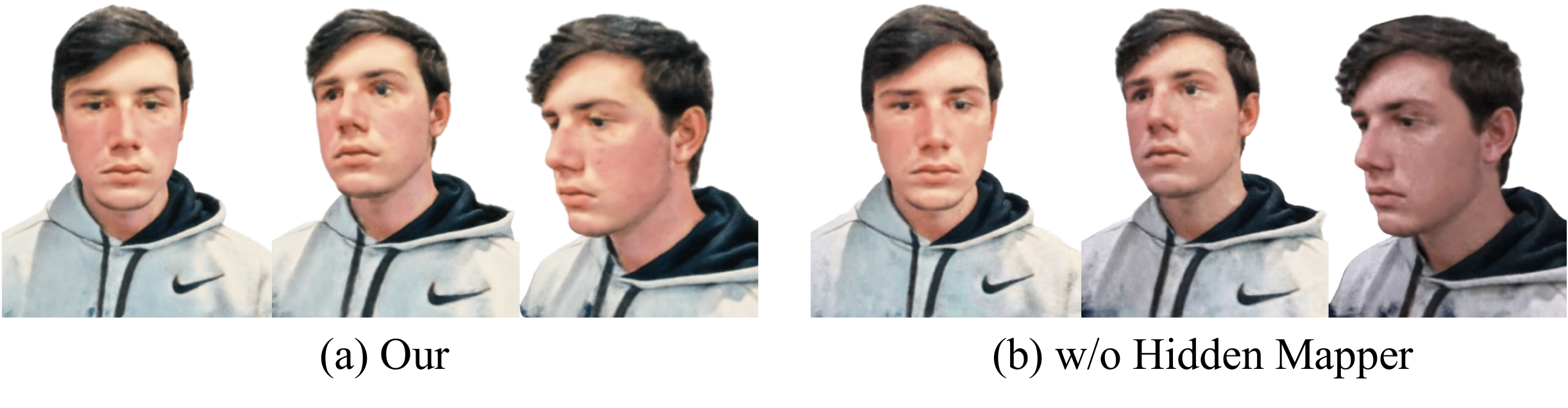}
     \vspace{-0.7cm}
     \caption{\textbf{Ablation study of hidden mapper.} (a) With hidden mapper, we can consistently transfer style to the entire scene. (b) W/o hidden mapper, the views outside the GAN training domain remain the original style and result in a scene with inconsistent styles across views.} 
     \label{fig:abl_hm}
\vspace{-0.2cm}
\end{figure}

We also show the importance of hidden mapper in real-world usage. In Fig.~\ref{fig:abl_hm}(b), as the color predicted by NeRF is conditioned on the viewing direction, without guided images on the GAN out-of-domain views, we will result in a scene with inconsistent styles from different views. 

\subsection{Applications}

\begin{figure}[t]
     \centering
     \includegraphics[width=1\linewidth]{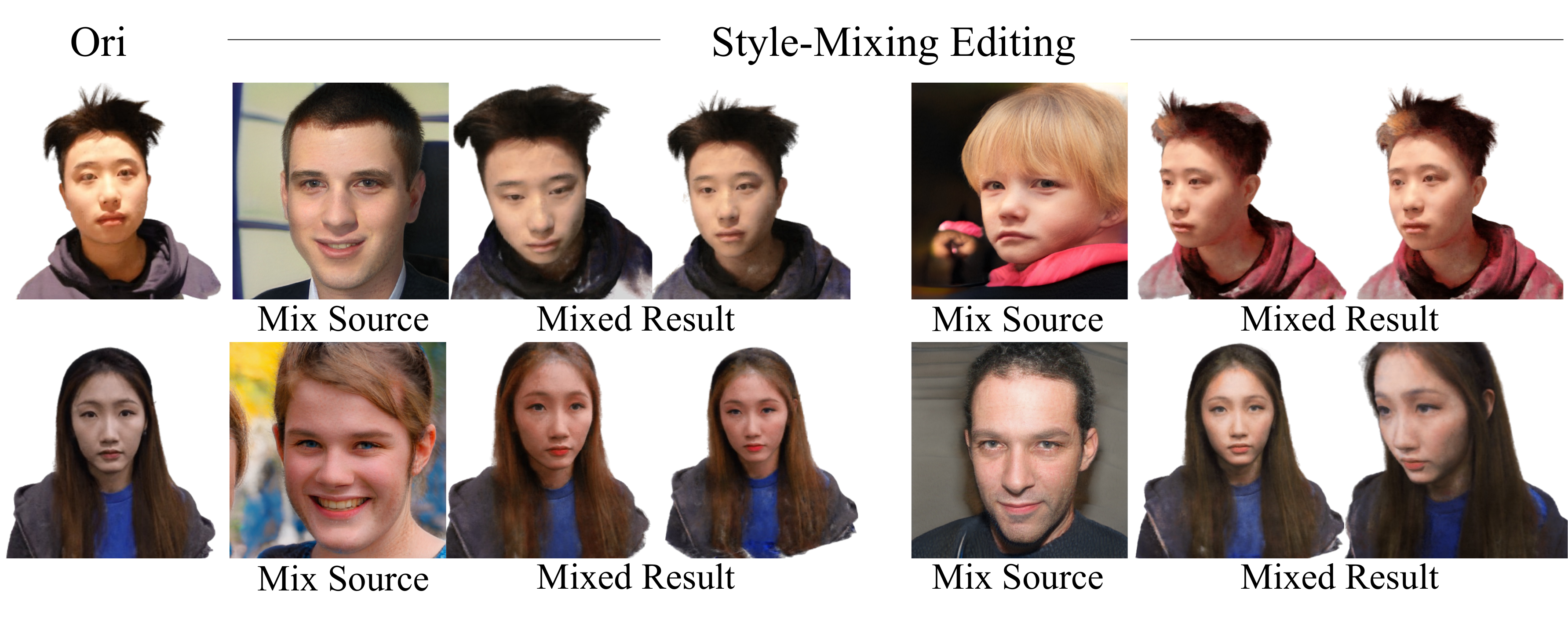}
     \caption{\textbf{Style-Mixing Result.} Our method can adapts Style-Mixing to 3D. Refer to the project page for 360$\degree$ results.}
     \label{fig:style_mix}
\end{figure}

\textbf{Style Mixing.} We can create the style-mixed image using StyleGAN. We apply style mixing~\cite{StyleGAN} to the frontal image latent code $\bar{w}$ and the $w_{style}$. Fig.~\ref{fig:style_mix} provides the result on various views, showing that our method can successfully transfer the mixed style from the source image, even with large appearance and environmental changes.

\begin{figure}[t]
     \centering
     \vspace{-0.2cm}
     \includegraphics[width=1\linewidth]{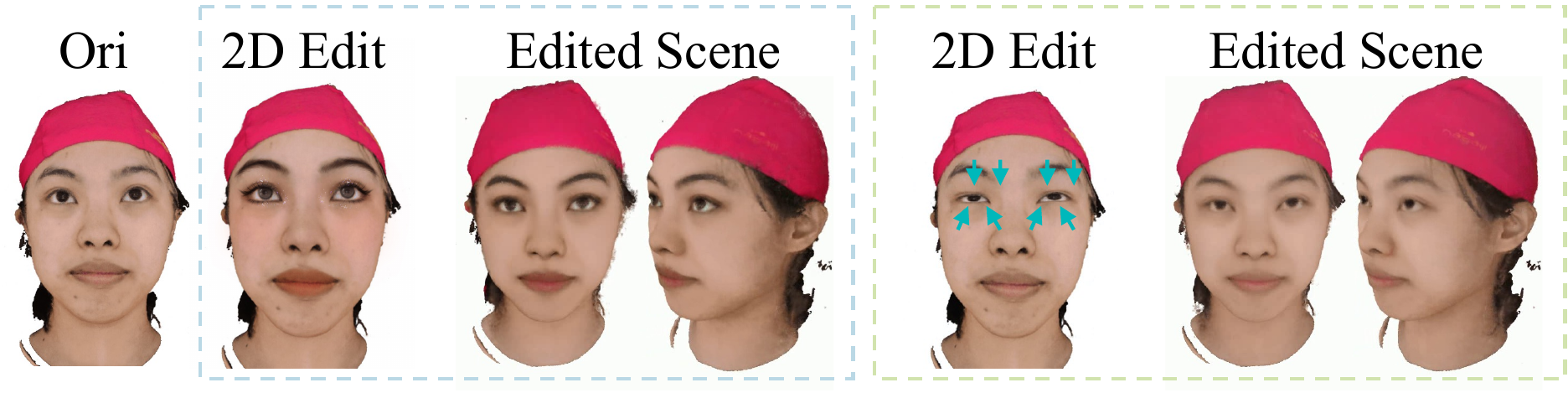}
     \caption{\textbf{Interactive Editing Result.} 360 $\degree$ results are in the project page. }
     \label{fig:ie_result}
\vspace{-0.2cm}
\end{figure}

\textbf{Interactive Editing.}
As our inversion techniques work well and our method can produce an identity-preserved image, we can edit the frontal image with any existing image editing tools (even non-learning-based tools such as photoshop!) and get the projected latent code $\bar{w}_{style}$. We can use our latent adjustor and hidden mapper to get stylized guided samples and perform 3D editing. 
In Fig.~\ref{fig:ie_result}, the left example is edited by makeup transfer tool and the right one is edited by image warping tool. 
Our method can successfully transfer both 2D editings to 3D. 
This indicates that our method is not limited by StyleGAN-based editing, but instead it can leverage any 2D image editing tool to edit one view of the 3D scene and generalize the effect to all views to produce an edited 3D scene.

\textbf{Class-agnostic 3D Style Transfer.}
3D style transfer aims to generate photorealistic images from arbitrary novel views given a style image. 
The major difficulty is how to achieve cross-view consistent style transfer. The StyleGAN model trained on wikiArt~\cite{ref:wikiArtRefined} can generate class-agnostic paintings. With this pre-trained model, we can perform class-agnostic 3D style transfer using the hidden mapper. For each image on the NeRF training set, we use the hidden mapper to map them into a common pre-trained StyleGAN hidden space and generate the latent code $w_{style}$ for the guided style image to perform style transfer. 

\begin{figure}[t]
     \centering
     \includegraphics[width=1\linewidth]{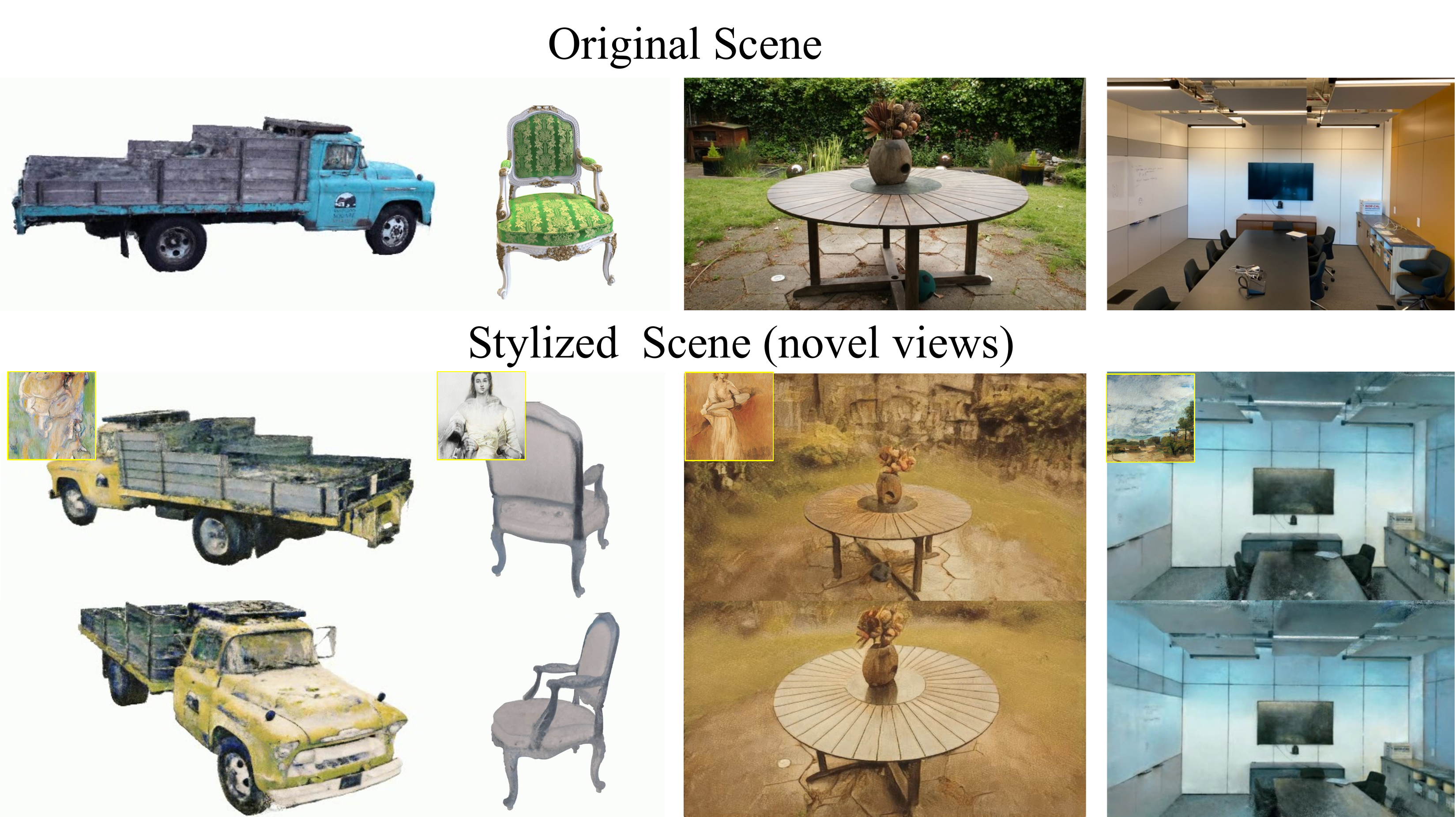}
     \vspace{-0.2cm}
     \caption{\textbf{3D Style Transfer.} The style-guided image is on the top-left corner of the stylized scene.}
     \label{fig:style_trans}
\vspace{-0.2cm}
\end{figure}

As depicted on Fig~\ref{fig:style_trans}, our method can successfully transfer the style from a given image without loss of view-consistency and works for any class. Our simple approach can achieve comparable results with the current state-of-the-art 3D style transfer methods.
By mapping the image into different StyleGAN blocks and changing the style-mixing factor, we can also enable the control of the abstract level and strength of stylization. The style transfer is also temporal consistent, combined with deformable NeRF will allow our method generalize to dynamic scenes. Details and more results can be found in the Appendix.

\section{Conclusion}

We present an efficient learning framework to bridge the gap between 2D editing and 3D editing, which is realized by two novel modules and a stable finetuning strategy. 
Our method can successfully transfer various editing patterns to 3D scenes, including text prompts, style-mixing, interactive warping, and style transfer. Our method outperforms all previous 3D editing methods with more flexible control and can support 360$\degree$ editing by overcoming the domain shift problem in 2D GAN and 3D-aware GAN using self-supervised training techniques. 

{\small
\bibliographystyle{ieee_fullname}
\bibliography{ref}
}

\appendix
\clearpage
\section{Conditional NeRF Architecture}

\begin{figure}[h]
     \centering
     \includegraphics[width=0.8\linewidth]{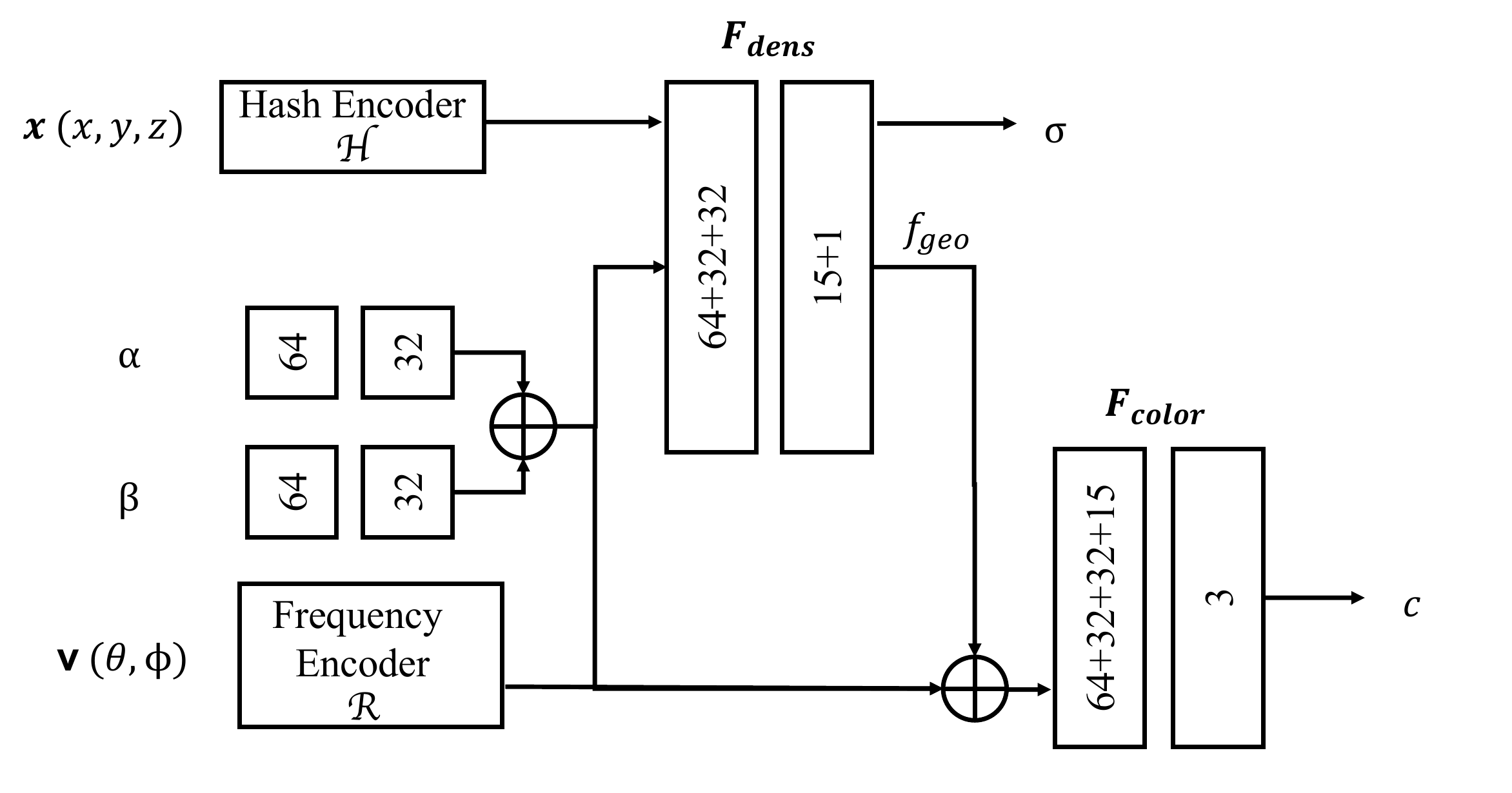}
     \caption{\textbf{Conditional NeRF Architecture.}}
     \label{fig:nerf_arc}
\end{figure}

Fig.~\ref{fig:nerf_arc} provides a detailed network architecture of the conditional NeRF. Compared to previous approaches~\cite{CLIPNeRF,EditNeRF} that train the network on massive 3D data, our network can be much lighter. This also helps us achieve faster speed in the style converting step.

\section{Differentiable Style Decomposition}

In the main paper (Sec. 3.2), we propose the differentiable style decomposition to decomposite the latent space $\mathcal{W}$ into an orthogonal basis $V$. 
Here we summarize the problem formulation and explicitly derive the gradients for back-propagation. 

We first sample a large batch of $z\in\mathcal{N}(0,\mathbf{I})$ (normal distribution), then use the mapping network to get a large set of styles $w\in\mathcal{W}$. Let $\mathbf{W} = \{w\} \in \mathbb{R}^{N\times D}$, we can calculate the covariance matrix as: 
\begin{equation}
    \boldsymbol{\Sigma} = \text{COV}(\mathbf{W}) = \frac{1}{N-1} (\mathbf{W}-\mathbf{\bar{W}})^\top (\mathbf{W}-\mathbf{\bar{W}})\,,
\end{equation}
where $\mathbf{\bar{W}} = \frac{1}{N}\sum_{i=1}^N \mathbf{W}_i$. 

\paragraph{Eigen-decomposition.} 
To explore disentangled styles in the latent space $\mathcal{W}$, we adopt the eigen-decomposition to find an orthogonal basis $V\in \mathbb{R}^{D\times k}$, consisting of eigenvectors w.r.t the top-$k$ eigenvalues. The formulation is 
\begin{align}
    \label{eq:eigen}
    \nonumber
    V \in \min_{U\in\mathbb{R}^{D\times k}} -\textbf{tr}(U^\top \boldsymbol{\Sigma} U) \\
    \text{subject to } U^\top U = I\,.
\end{align}
To solve Eq.~(\ref{eq:eigen}), we use a widely-used power iteration (PI) method~\cite{ref:eigen}, which is a numerical method to iteratively update until converge to eigenvectors. 

\paragraph{Backward Gradient Derivation.} To integrate Eq.~(\ref{eq:eigen}) in deep neural networks for end-to-end training, a straightforward way is to perform auto-differentiation and backpropagate through the power iterations. However, this method is in-efficient in both memory and computation. Thereby, we adopt the DDNs technique~\cite{Gould2019DeepDN} to explicitly derive the gradients for efficient backward propagation. Formally, we define $f(\boldsymbol{\Sigma}, V) = -\textbf{tr}(V^\top \boldsymbol{\Sigma} V)$ and $h(V) = V^\top V - I$. According to \cite{Gould2019DeepDN}, the gradient of $V$ w.r.t $\boldsymbol{\Sigma}$ can be calculated as: 
\begin{equation}
    \label{eq:ddn}
    \small
    DV(\boldsymbol{\Sigma}) = H^{-1}A^\top (A^\top H^{-1} A)^{-1} AH^{-1}B - H^{-1}B\,,
\end{equation}
where 
\begin{align}
    A & = D_{V}h(V),\\
    B & = D_{\boldsymbol{\Sigma} V}^2 f(\boldsymbol{\Sigma}, V)\,, \\
    H & = D_{V V}^2 f(\boldsymbol{\Sigma}, V) - \lambda D_{V V}^2 h(V)\,,
\end{align}
and $\lambda$ satisfies $\lambda^\top A = D_{V}f(\boldsymbol{\Sigma}, V)$. 

At neural network training time, the gradients flow from loss function to $V$, then flow through $\boldsymbol{\Sigma}$ according Eq.~\ref{eq:ddn}.

\section{Hidden Mapper Architecture}
\begin{figure}[h]
     \centering
     \includegraphics[width=1\linewidth]{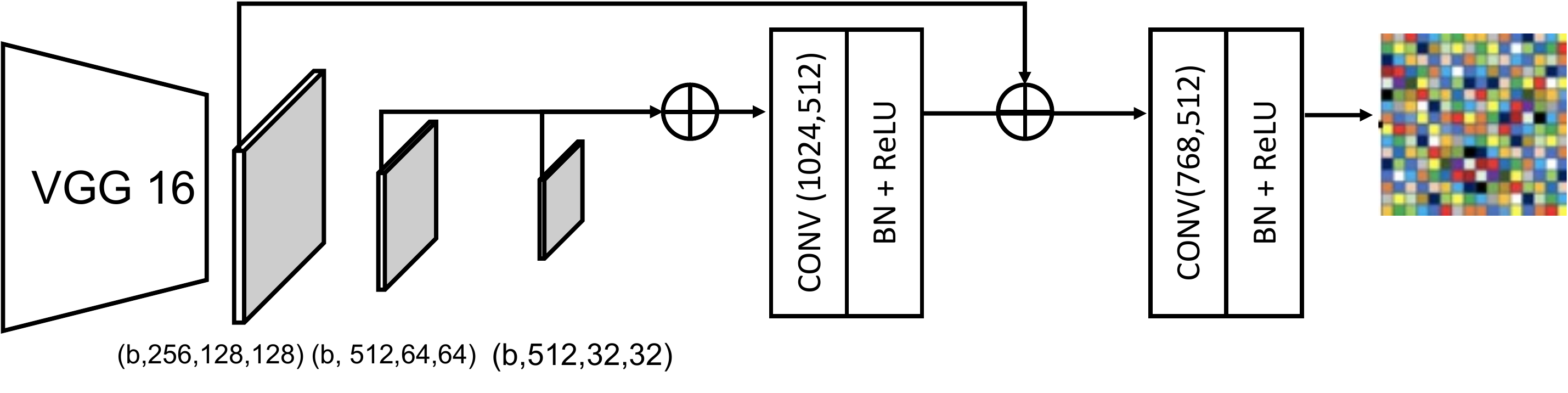}
     \caption{\textbf{Architecture of Hidden Mapper.}} 
     \label{fig:hidden_mapper_arc}
\end{figure}

Fig.~\ref{fig:hidden_mapper_arc} provides the hidden mapper architecture, which contains a pertained VGG16 as the backbone and a few convolutional layers to produce different sizes of StyleGAN hidden features maps ($F_{32 \times 32}$, $F_{64 \times 64}$, and $F_{128 \times 128}$). During self-supervised training, we randomly sample one of the three hidden maps to perform style-mixing. The training takes 20,000 iterations with a batch size of 2, which only performs once and can be used for all scenes. 

\section{3D Style Transfer}

\begin{figure}[h]
     \centering
     \includegraphics[width=1\linewidth]{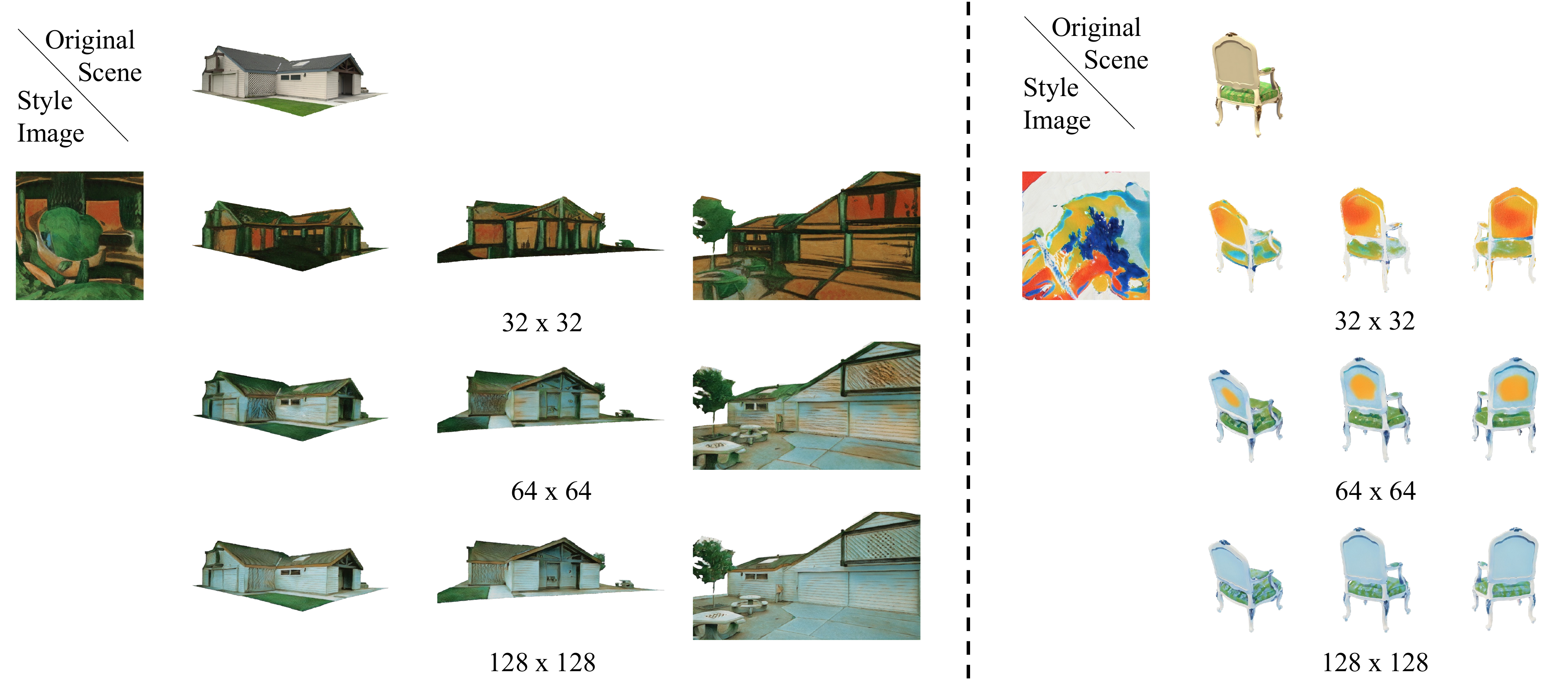}
     \caption{\textbf{Style Transfer Result on Different Level.}}
     \label{fig:style_level}
\end{figure}

Fig.~\ref{fig:style_level} further shows how our method can control the abstraction level of style transfer. Mapping the image to a shallow style block of StyleGAN will allow the transferred image more similar to the abstraction level of the style image. As we map the image to the hidden space, the shallow level will produce more abstract transferred results. 
If we map the image to the deeper layer, as shown in Fig.~\ref{fig:style_level} bottom row, the output images almost only share color information of the given style image. To control the strength of the style transfer, we can choose one image from the NeRF dataset, and then calculate its corresponding latent code. When performing style transfer, we can mix it with the $w_{style}$ with different mixing factors to control the style transfer strength. 

\begin{figure}[h]
     \centering
     \includegraphics[width=1\linewidth]{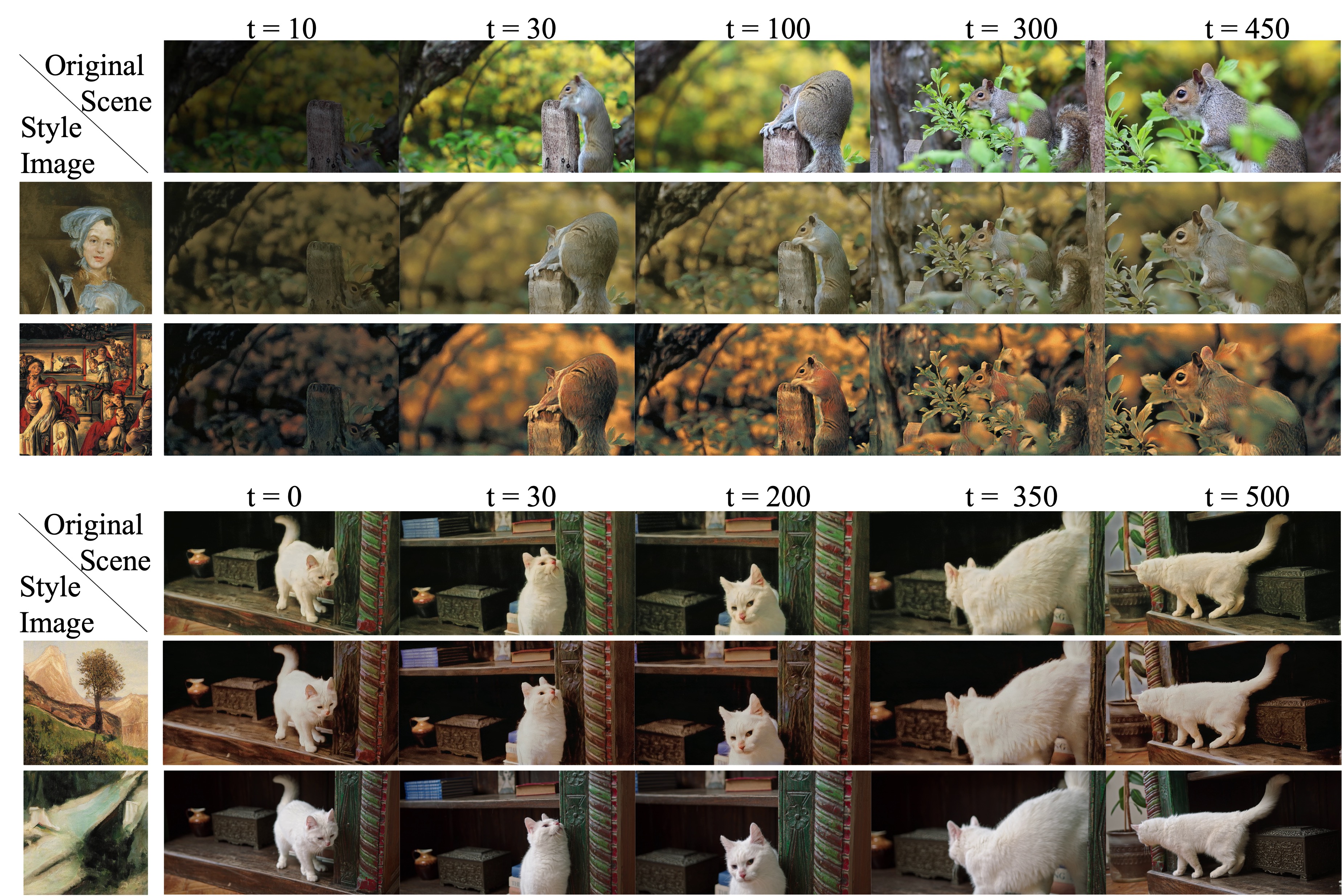}
     \caption{\textbf{Style Transfer Result on Video.}}
     \label{fig:style_video}
\end{figure}

Fig.~\ref{fig:style_video} further shows that our method can produce temporally consistent style transfer. The top example shows in t = 300, the style is still consistent even the camera changes its position.  The bottom examples show that for a fast-moving target, the style can also transfer consistently. This support that our method can even be applied to video style transfer and extend to deformable NeRF for 4D scene style transfer.

\section{Dataset Collection}

We capture the 360 $\degree$ video using a single camera around each participant and require the participant to sit and stay for one minute.  To make the background static but still under natural lightering conditions, we choose empty rooms for data collection. For each participant, we capture three videos. For each video, we first filter blurry frames by calculating the sharpness of each frame, and then we keep ~100 frames for each video. After filtering, we use COLMAP~\cite{Schnberger2016StructurefromMotionR} to compute the pose for each image and the camera intrinsics. Segmenting the background can enable faster training and reduce floating noise. We use the previous state-of-the-art video matting model~\cite{Sun2021MODNetVIP} to segment the foreground and save the foreground mask for each frame. After that, we use the instantNeRF~\cite{Mller2022InstantNG} to reconstruct each scene. For each participant, we kept the data sample with the best visual quality. 

\end{document}